\documentclass[conference]{IEEEtran}
\IEEEoverridecommandlockouts
\usepackage{cite}
\usepackage{amsmath,amssymb,amsfonts}
\usepackage{algorithmic}
\usepackage{graphicx}
\usepackage{textcomp}
\usepackage{xcolor}
\usepackage{epstopdf}

\def\BibTeX{{\rm B\kern-.05em{\sc i\kern-.025em b}\kern-.08em
    T\kern-.1667em\lower.7ex\hbox{E}\kern-.125emX}}
\begin{document}

\title{Automatic Detection of Cardiac Chambers Using an Attention-based YOLOv4 Framework from Four-chamber View of Fetal Echocardiography\\
}

\author{\IEEEauthorblockN{Sibo Qiao\dag, Shanchen Pang\dag, Gang Luo\ddag, Silin Pan\ddag, Xun Wang\dag, Min Wang\dag, Xue Zhai\dag, Taotao Chen\S
}
\IEEEauthorblockA{\textit{\dag College of Computer Science and Technology, China University of Petroleum}\\
\textit{\ddag Heart Center, Qingdao Women and Children's Hospital}\\
\textit{\S Department of ultrasound, Qingdao Women and Children's Hospital}}\\
}

\maketitle

\begin{abstract}
Echocardiography is a powerful prenatal examination tool for early diagnosis of fetal congenital heart diseases (CHDs). The four-chamber (FC) view is a crucial and easily accessible ultrasound (US) image among echocardiography images. Automatic analysis of FC views contributes significantly to the early diagnosis of CHDs. The first step to automatically analyze fetal FC views is locating the fetal four crucial chambers of heart in a US image. However, it is a greatly challenging task due to several key factors, such as numerous speckles in US images, the fetal cardiac chambers with small size and unfixed positions, and category indistinction caused by the similarity of cardiac chambers. These factors hinder the process of capturing robust and discriminative features, hence destroying fetal cardiac anatomical chambers precise localization. Therefore, we first propose a multistage residual hybrid attention module (MRHAM) to improve the feature learning. Then, we present an improved YOLOv4 detection model, namely MRHAM-YOLOv4-Slim. Specially, the residual identity mapping is replaced with the MRHAM in the backbone of MRHAM-YOLOv4-Slim, accurately locating the four important chambers in fetal FC views. Extensive experiments demonstrate that our proposed method outperforms current state-of-the-art, including the precision of 0.919, the recall of 0.971, the F1 score of 0.944, the mAP of 0.953, and the frames per second (FPS) of 43.
\end{abstract}


\section{Introduction}
\label{sec:introduction}
\IEEEPARstart{R}{ecently}, echocardiography is generally utilized in prenatal fetal heart malformation screening for its advantages of relative security, high sensitivity, noninvasive property and real-time imaging. Specially, echocardiography has been considered as an important screening tool for early diagnosis of fetal congenital heart diseases (CHDs). A four-chamber (FC) view is an important ultrasound (US) plane used in prenatal diagnosis of CHDs. The FC view can clearly depict the morphology and size of the four chambers of heart, which can be adopted to comprehensively evaluate the development of fetal heart, such as left cardiac dysplasia or right cardiac dysplasia \cite{1,2,3}. In order to accurately evaluate the fetal heart, the primary step is to detect the location and identify the categories of the fetal four chambers in FC views. However, the FC planes containing the four cardiac chambers are captured by different operators at various views, which makes them may be greatly diverse and illegible. Therefore, it may be very challenging for an inexperienced obstetrician to distinguish the four chambers. Moreover, training a skillful obstetrician for diagnosis of CHDs can be costly and consume significant amounts of time \cite{4}. 

To alleviate above problems, a computer-aided method, helping obstetricians to locate the four chambers automatically, has attracted much attention in recent years \cite{6,7,8,9,10,46}. The computerized procedures can assist obstetricians to improve the efficiency of diagnosis of CHDs with automatic mechanism, and shorten the learning curve for several novices. However, there are still several challenges to identify the fetal chambers of heart in automation. First, fetal FC planes with lower resolution and more speckles make it difficult to extract distinguished features of the four cardiac structures; Then, the fetal cardiac structures occupy a small portion of a US image, and present various appearance in different FC planes due to relative movement of the probe and fetus; Finally, the physical boundary between the four cardiac structures is not distinct or even disappeared in FC views when the mitral valve, tricuspid valve, atrium or ventricle are opening. In this case, there is a high degree of similarity among the four cardiac structures, which the identification of the structures highly depends on obstetrician's extensive experience. Considering the above challenges, an effectively automatic fetal cardiac structures detection system desires to gain features that are context-invariant, position-insensitive and structure-specific.

Deep learning has recently been widely used in various fields of medical images due to its strong ability to learn invariant features, such as breast cancer screening \cite{11,12}, tumor classification \cite{13} and the diagnosis of brain disease \cite{14}. At present, deep learning is also increasingly popular in heart US images for the classification, detection and segmentation of the adult heart \cite{15} and fetal heart \cite{17,18,19}. Convolutional neural network (CNN) is one of the representative methods of deep learning, which has strong feature power and learns robust and discriminative features from medical images. The invariable features of fetal heart learned by CNN can help automatic detection system to locate fetal cardiac structures accurately. However, the fetal heart actually has small scale with relatively ambiguous anatomical structural appearance, making it significantly difficult to locate fetal heart than adult heart in a US image. Moreover, several adult heart detecting methods are not appropriate in fetal heart since the position of fetal heart changes constantly in utero. Furthermore, most of exiting CNN methods learn heart features from a whole image. An image-level feature contains not only specific heart but also background clutter and occlusion, which brings several troubles to precisely locate cardiac structures.

The visual attention mechanism has shown its superiority in several computer vision tasks such as image classification \cite{20}, detection \cite{21,22}, caption \cite{23,24} and segmentation \cite{25}, which mainly owes to the property as similar with the human visual system. Attention mechanism, as the human visual system, captures a series of partial glimpses and is interested in highlighted parts selectively to explore visual context, but not processes a whole image at once. The attention module guides automatic detection system to learn from the regions with specific information in US images, extracting purer features of fetal cardiac structures. Hence, we propose a multistage residual hybrid attention module (MRHAM), which focuses more on spatial and content information of fetal cardiac structures in US images.

The main contributions of this paper are as follows for the localization of four fetal cardiac anatomical structures in US images:

(1) We propose an attention module namely MRHAM. The MRHAM mainly adopts convolutional operation to focus on what is meaningful and where is emphasis simultaneously in a given feature map, integrating residual identity mapping to alleviate the problem of information loss.

(2) In order to improve the ability of learning invariable features, the CSPDarknet53 as backbone of YOLOv4 \cite{27} adopts MRHAM instead of the original residual mapping module.

(3) We reduce the number of MRHAM in CSPDarknet53 for relieving overfitting and named it MRHAM-CSPDarknet53-Slim. MRHAM-CSPDarknet53-Slim is shallower 16-layers than CSPDarknet53, which significantly decreases the complexity of the whole object detection system. The improved YOLOv4 is named as MRHAM-YOLOv4-Slim.

The rest of this paper is organized as follows: In Section II, we provide several related literatures on YOLO model, attention mechanism in computer vision and deep learning in fetal US images. In Section III, we introduce our proposed framework for detecting fetal four cardiac chambers. In Section IV, we conduct extensive evaluations on our proposed detection system with two public benchmarks and the fetal cardiac US dataset. In Section V, we present the conclusion of this paper and discuss the future plans for CHDs.

\section{Related Work}
In this section, we begin by briefly providing an overview of YOLO object detection model. Then, we introduce the effectiveness of attention mechanism in computer vision tasks and finally describe the application of deep learning in echocardiography.
\subsection{YOLO object detection model}
Redmon J et al. \cite{28} first propose the YOLO object detection model, which frames object detection as a special regression problem, identifying the object and capturing a good bounding box (BBox) simultaneously in a single step. Although the accuracy of YOLO is not as high as that of Fast R-CNN \cite{29} and Faster R-CNN \cite{30}, it is considerable superior to these two models in terms of detection speed.  This is particularly important for detection tasks with high real-time requirements.

Then, YOLOv2 \cite{31} and YOLOv3 \cite{32} are proposed to improve the detection performance. The YOLOv2 runs k-means clustering on the training set BBoxes to automatically find 9 precise anchors. The experimental results demonstrate that the YOLOv2 with cluster priors outperforms using hand-picked anchors. The YOLOv3 employs Darknet53 as a new CNN backbone to extract features from input images. The Darknet53 network adopts convolutional operation without pooling method in its architecture. Hence, the Darknet53 shows a better performance and more speed than other backbones (e.g., ResNet50 \cite{45} and ResNet101 \cite{45}).

It is common knowledge that an object detection network has more computationally expensive than an ordinary CNN. Moreover, training an excellent object detection model requires expensive physical hardware. Most of the object detection models show the best performance only with the GPU equipment with strong computing power. To alleviate this problem, Bochkovskiy et al. \cite{27} propose a powerful detection model, YOLOv4, which integrates several new approaches in various fields. We can get a fast and accurate YOLOv4, and for which training requires only one 1080Ti GPU or 2080Ti GPU. YOLOv4 has a good performance on several large public datasets. However, due to the lack of fetal cardiac US images as well as US images with lower resolution and more speckle noise in this research, the YOLOv4 with expensive computation is prone to overfitting and hard to accurately capture the fetal cardiac structures in US images, resulting in poor performance in the localization fetal heart task. Hence, we must make several corresponding improvements to YOLOv4 model according to the characteristics of fetal cardiac dataset so that the model can achieve the best performance.
\subsection{Attention mechanism in computer vision}
Attention plays an extremely important role in a human perceptive system \cite{34}. An important property of the human perceptive system is that one does not attempt to capture a full visual scene at once. Instead, human perceptive system selectively focuses on highlights of the scene, taking advantage of a set of local features captured to better understand the entire visual structure \cite{35}. The process of the detection is very similar to the human perception while there are many obstacles in object detection, such as object occlusion, overlapping and feature misalignment. Moreover, a detection system is difficult to focus on target objects when given input images with lower resolution and more background clutter. BBoxes predicted by a detection system may localize arbitrary contours of various objects, which reduces the performance of the system.

To solve above problems, several networks adopt attention to learn key features of target objects. Recently, attention mechanism has demonstrated its effectiveness in many tasks, including image classification [20], object detection [21-22] and image caption \cite{23,24}. Fan et al. \cite{22} improve the region proposal network (RPN) in the Faster R-CNN \cite{30}, where attention is added in the network. The RPN with attention precisely filters out the BBoxes containing only background as well as none support categories. Hu et al. \cite{36} focus on the inter-channel relationship of feature maps, proposing a squeeze-and-excitation (SE) block. A collection of SE blocks is stacked simply to construct an SE network which achieves better classification performance on the ImageNet dataset. When extracting features from CNN, we should pay attention to not only interactions between channels of feature maps, but also their spatial information. Therefore, Woo et al. \cite{37} integrate channel and spatial information of feature maps, presenting a convolutional block attention module (CBAM). However, the CBAM mainly adopts pooling operation to process feature maps, which is not fitting into Darknet's style. Hence, we propose MRHAM, a new attention module inspired by CBAM \cite{37}. The MRHAM primarily adopts convolutional operation to learn contextual information from feature maps. Moreover, the residual structure is employed in the MRHAM to improve the performance.
\subsection{Deep learning in fetal US images}
Deep learning especially CNN is gradually applied in localizing the fetal heart in US images. Oktay et al. \cite{15} present an anatomically constrained neural network (ACNN) to detect left ventricle endocardium in adult US images. The ACNN integrates prior information including the scale and label of heart, which automatically learn more meaningful boundary knowledge of left heart. To precisely locate left ventricle in adult US images, Leclerc et al. \cite{7} propose a neural network with Encode Decode structure. The network recognizes the left ventricle in pixel-level more accurately. Recently, the location and recognition of fetal heart in US images has gradually attracted several interests in academics. Christopher et al. \cite{17} early attempt to describe the fetal cardiac morphology automatically in echocardiography. They train an adaptive regression random forest to predict the position of the fetal heart in US images. Maraci et al. \cite{18} propose a framework for automatic detection of fetal heart. The framework first screens US images containing fetal heart from echocardiography videos. Then, they use a conditional random field model to find the location of fetal heart in US images.

Recent studies show that CNN is more suitable for recognizing patterns in US images. Compared with traditional machine learning methods, CNN has strong power to explore contextual information and the ability to learn representations from an entire image. Sundaresan et al. \cite{8} propose a full CNN model to detect the location of fetal heart in a FC view. Xu et al. \cite{9} propose a dilated convolution chain model, which adopts dilated convolution to expand the receptive field, further extracting abundant fetal heart information. The chain model integrates the global and local information of feature maps, more precisely locating fetal cardiac structures in FC planes. Patra et al. \cite{10} employ Faster R-CNN to extract the region of interest (ROI) of fetal heart in FC views, and then use the global spatial information of ROI to detect the anatomical structures. To improve the detection speed, Dong et al. \cite{46} propose a new Aggregated Residual Visual Block Network (ARVBNet) to detect key anatomical structures of fetal heart. The ARVBNet is a one-stage detector, which ARVB blocks are embbedded into the top layers of the SSD network \cite{47}.

Inspired by previous works, we still adopt deep learning methods to explore the problem of localization for four fetal cardiac chambers automatically in US images. To further import detection accuracy and speed, we introduce a new object detector. Next, we will provide details of our proposed model.

%The author names and affiliations could be formatted in two ways:
%\begin{enumerate}[(1)]
%\item Group the authors per affiliation.
%\item Use footnotes to indicate the affiliations.
%\end{enumerate}
%See the front matter of this document for examples. You are recommended to conform your choice to the journal you are submitting to.

\section{Our proposed method}

To solve the localization problem of fetal four cardiac chambers (e.g., RA, RV, LA, LV) in US images, we propose an improved YOLOv4 model named MRHAM-YOLOv4-Slim. The CSPDarknet53 as a backbone of YOLOv4 is primary used to extract features from input US images for detection tasks. The extracted features with strong distinguishable ability can improve the performance of a detector. Hence, we make several improvements on the CSPDarknet53. In the following, we begin by providing details for CSPDarknet53 architecture. Then, we detail the proposed MRHAM-YOLOv4-Slim architecture.
\subsection{CSPDarknet53}
CSPDarknet53 is a backbone of YOLOv4 model, which is a new 67-layers deep CNN proposed by Alexey et al. \cite{27}. The CSPDarknet53 adopts convolution instead of pooling completely, where the scale of output feature maps and receptive filed are controlled by adjusting strides. CSPDarknet53 introduces Mosaic, a new data augmentation method while processing images, which mixes 4 input images by random scaling, random cropping and random arrangement. The augmentation method enriches training dataset by increasing the diversity of samples, improving the generalization capability of the backbone. Moreover, batch normalization calculates activations from the mixed image on each layer, which considerably reduces the dependencies for a large mini-batch size. That is why a conventional GPU is used to train YOLOv4 model achieving good performance.

CSPDarknet53 employs a cross stage partial (CSP) structure which adds a new skipping connection to the original residual identity mapping, merging valuable information between low-levels and high-levels. In the training phase, the CSP structure uses gradients repeatedly, reducing the computational complexity of the whole model. The Mish activation function instead of Leaky ReLU is adopted in CSPDarknet53, which enables gradients to flow smoothly in the backbone architecture. Similar to Dropout used in a fully connected network, Dropblock regularizes a CNN to address the overfitting. The CSPDarknet53 employs the Dropblock to randomly select an area of a feature map and set it to zero.

\subsection{MRHAM-YOLOv4-Slim}
YOLOv4 has more expensive computations, which can achieve significant performance in the case of abundant samples (e.g., COCO dataset). However, the fetal cardiac dataset used in this paper is not enough which consisting of 1,250 US images. if the YOLOv4 with high-complexity is still utilized to fit fetal cardiac US images, resulting in a considerable decline in the performance of the model.

To mitigate above problems, we present an improved object detector namely MRHAM-YOLOv4-Slim, as illustrated in Fig. 1. An input image is fed into the model, generating three feature maps with different scales. Each feature map is responsible for detecting targets of various scales. We empirically argue that the purpose of adopting more residual modules in CSPDarknet53 is to extract greatly discriminative features without the problem of vanishing/exploding gradients in a large dataset. Therefore, to better fit the small-scale fetal cardiac US dataset in this article, we make several modifications on CSP architecture in the CSPDarknet53. Specifically, we first reduce the number of residual identity mappings in the CSP, where the residual identity mapping in the 3rd, 4th and 5th CSP of CSPDarknet53 is reduced from 8, 8 and 4 to 4, 4 and 2, respectively. Empirically, the shallower the model, the more difficult it is to extract more effective representations. As described in Fig. 1, to improve the performance of the backbone, we then propose a MRHAM to replace the residual identity mapping in the CSP architecture. Hence, the improved CSPDarknet53 trained with small-scale fetal medical dataset still has better performance. The improved CSPDarknet53 model is named as MRHAM-CSPDarknet53-Slim which decreases by 16 layers. To evaluate MRHAM-CSPDarknet53-Slim, we conduct extensive experiments on two public datasets, and the experimental results are shown in Table I.

\begin{figure*}
	\centering
		\includegraphics[width=0.57\textwidth]{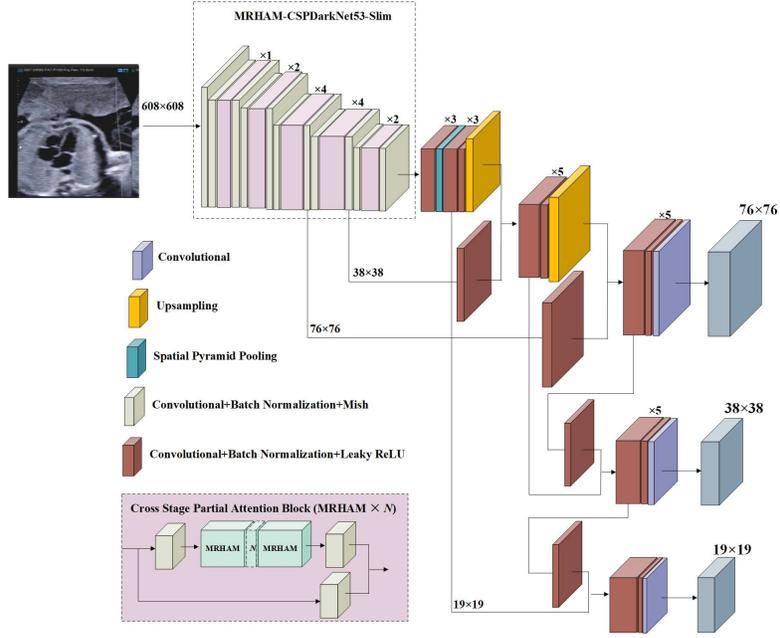}
	\caption{The architecture of MRHAM-YOLOv4-Slim model. MRHAM $\times$ N denotes the number of MRHAM is adopted in the model. An input image is fed into the model to generate three feature maps with different scales, where the feature maps of $76 \times 76$ are used to locate four chambers with small scales, the feature maps of $38 \times 38$ are used to locate four chambers with medium scales, and the feature maps of $19 \times 19$ are used to locate four chambers with large scales.}
	\label{FIG:1}
\end{figure*}

As described in Fig. 1, similar to YOLOv4, MRHAM-YOLOv4-Slim also adopts several special structures such as Spatial Pyramid Pooling (SPP) \cite{38}, Feature Pyramid Network (FPN) \cite{33} and Path Aggregation Network (PAN) \cite{39}. The SPP module is utilized to increase the variability of the receptive field, connecting feature maps with different scales to increase information diversity. The structure of FPN extracts in-network feature hierarchy with a top-down path, which augments to propagate semantically excellent features. Furthermore, the structure of PAN adopts a bottom-up path to enhance feature pyramid with precise localization patterns existing in low-levels. The FPN and PAN are used together to further improve the localization ability of the full feature maps hierarchy by propagating strong signals of low-level information.

As showed in Fig. 1, when predicting objects locations and categories, MRHAM-YOLOv4-Slim transmits the extracted feature maps to three different branches to obtain three feature grid maps with various scales for detecting objects of different sizes. Generally, a detector will generate many bounding boxes for each object. In order to effectively filter out redundant BBoxes for each object, MRHAM-YOLOv4-Slim employs the DIOU NMS \cite{40} which considers the distance between the center points of two BBoxes. When adjusting the coordinates of the bounding boxes, MRHAM-YOLOv4-Slim adopts the CIOU loss \cite{40} which simultaneously considers the overlapping area between BBoxes and anchor boxes, the distance between center points, and the aspect ratio. CIOU loss \cite{40} enables the model to achieve better convergence speed and accuracy on the BBoxes regression problem.
\subsection{MRHAM}
The attention mechanism is integrated into the backbone to focus more on ROI of feature maps, extracting crucial features related to fetal cardiac structures simultaneously ignoring the irrelevant features, e.g., background clutter. Hence, we present a MRHAM architecture including a residual channel attention (RCA) module and a residual spatial attention (RSA) module, which considers spatial context and channel interactions simultaneously, as shown in Fig. 2.
\begin{figure}[!htbp]
	\centering
		\includegraphics[width=0.45\textwidth]{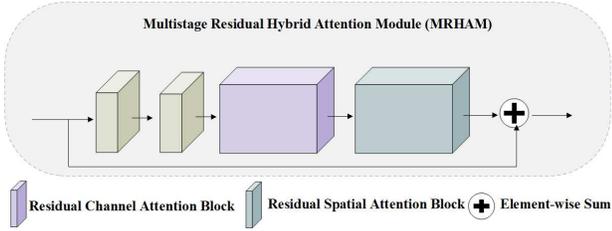}
	\caption{Multistage residual hybrid attention module.}
	\label{FIG:2}
\end{figure}

From Fig. 2, the MRHAM architecture is constructed by simply stacking the RCA module and the RSA module in a sequential order. Given an input feature map, the MRHAM architecture learns meaningful information about a target object in the feature map as well as the location of a target. In fact, although the MRHAM architecture does not increase the depth of a whole model, it expands the width of the entire architecture, which is of great help in improving the performance of the model. Next, we describe several details for the RCA and the RSA, respectively.

\textbf{Residual channel attention block.} To explore inter-channel interactions of feature maps, we propose a residual channel attention block, as shown in Fig. 3.
\begin{figure}[!htbp]
	\centering
		\includegraphics[width=0.45\textwidth]{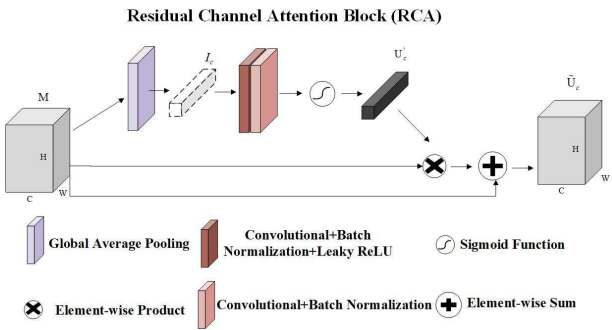}
	\caption{Residual channel attention block.}
	\label{FIG:3}
\end{figure}
%Here are two sample references: \cite{Feynman1963118,Dirac1953888}.

The RCA block is adopted to focus on valuable information of a feature map, extracting distinctive features of a target object. As known to us, the receptive field is limited in conventional convolution operation, where the filters only capture a certain local area in a feature map. As illustrated in Fig. 3, we first employ global average pooling operation to explore effective contextual information of a full feature map in the RCA block, which an input feature map $M \in R^{H\times W \times C}$ is aggregated to generate a channel descriptor $I_{c} \in R^{1 \times 1 \times C}$. From considering of implementation aspects, a channel descriptor $I_{c} \in R^{1 \times 1 \times C}$ is generated by squeezing the spatial dimensions of $M \in R^{H\times W \times C}$ , e.g., the c-th element of $I_{c}$ is computed by:
\begin{equation}
i_c = \frac{1}{H \times W} \sum_{i=1}^{H} \sum_{j=1}^{W} m_c(i,j)
\end{equation}
where $I_c = [i_1, i_2, ..., i_c], M = [m_1, m_2, ..., m_c]$

Then, to further explore the information of $I_c$, we utilize a 2-layers CNN to capture inter-channel nonlinear relationship. As in \cite{36,37}, we set the 1st layer of the 2-layers CNN as a reduction layer where reduction ratio is set to 16 to reduce parameter overhead. Moreover, we adopt a simple gating mechanism to process channel descriptor after the CNN, generating a logical channel information structure $U_c^{'}$. Furthermore, we merge the logical spatial structure $U_c^{'}$ and the input feature map $M$ using a weighted sum operation to generate a temporary channel attention map. The above attention process is calculated as:
\begin{equation}
U_c^{'} = \sigma(W_2 \phi (W_1I))
\end{equation}
\begin{equation}
U_c = U_c^{'} \times M
\end{equation}
where $\phi(.)$ denotes Leaky ReLU activation function; $W_1 \in R^{\frac{C}{r} \times C}$ and $W_2 \in R^{C \times \frac{C}{r}}$; $\sigma(.)$ refers to sigmoid activation function.

Finally, in order to mitigate the loss of input maps information, we conduct an element-wise summation operation on the input feature map and the temporary channel attention map through a residual connection to obtain the final channel attention feature map $\widetilde{U}$, which is computed as:
\begin{equation}
\widetilde{U} = U_c + M
\end{equation}

\textbf{Residual spatial attention block.} In order to explore the spatial relationship between local regions of a feature map, we propose a residual spatial attention block, as shown in Fig. 4.
\begin{figure}[!htbp]
	\centering
		\includegraphics[width=0.45\textwidth]{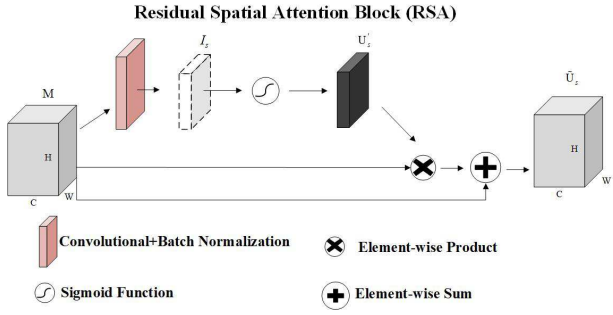}
	\caption{Residual spatial attention block.}
	\label{FIG:4}
\end{figure}

Different from the RCA block, the RSA block focuses on the location of the target object in the input feature maps. Similarly, to follow the CSPDarknets style, we adopt convolutional methods to capture a spatial context descriptor. As described in Fig. 4, we first employ a $7 \times 7$ convolutional layer to process the input feature maps, where merges inter-spatial relationship of maps to produce a spatial descriptor $I_s \in R^{H \times W \times 1}$. Then, we still gate the spatial descriptor with a sigmoid activation function, which generates a logical spatial information structure $U_s^{'}$. Furthermore, similar to the RCA, the logical spatial information structure is carried out a weighted sum operation with the input feature maps, which produces a temporary spatial attention map $U_s \in R^{H \times W \times C}$. The above attention process is computed as:
\begin{equation}
U_s^{'}(M) = \sigma(Conv^{7 \times 7}(M))) = \sigma(I_s)
\end{equation}
\begin{equation}
U_s = U_s^{'}(M)
\end{equation}
where $\sigma(.)$ refers to sigmoid activation function; $Conv^{7 \times 7}(.)$ denotes a $7 \times 7$ convolutional layer.

Similar to the RCA, we eventually implement an element-wise summation operation on two crucial maps, namely the input feature map and the temporary spatial  attention map, through a residual connection to obtain the final spatial attention feature map $\widetilde{U} \in R^{H \times W \times C}$, which encodes where to highlight or suppress. The computational process is as follows:
\begin{equation}
\widetilde{U}_c = U_c + M
\end{equation}

\section{Experiments}
\subsection{Datasets and Evaluation Measures}
The fetal echocardiography dataset used in this work is obtained from the Qingdao Women and Children's Hospital, which is a grade 3 A hospital mainly for the treatment of gynecological and childhood diseases. Due to its clinical authority, Qingdao Women and Children's Hospital collects extensive relevant US images of fetal heart diseases from all over china, including CHDs. The experimental dataset of random selection contains 1,250 FC views obtained from the pregnant US examinations at various scanning angles, conducting on 1,000 healthy pregnant women by different doctors. There are varying degrees of artifacts, speckle noise and fuzzy boundaries in the dataset, making it very suitable for verifying the effectiveness of the MRHAM-YOLOv4-Slim in dealing with the above detection tasks. All images are labeled by two experienced radiologists, and the annotated images undergo a rigorous verification.

To evaluate our proposed MRHAM module, we adopt two openly available benchmarks namely CIFAR-10 \cite{26} and CIFAR-100 \cite{26}. CIFAR-10 \cite{26} dataset is composed of 6,000 natural images of size $32 \times 32$ of 10 categories. CIFAR-100 \cite{26} dataset consists of 6,000 natural images of size $32 \times 32$ of 100 categories. The datasets of CIFAR-10 \cite{26} and CIFAR-100 \cite{26} are used to test backbones with or without attention mechanism.

In addition to these datasets that are used in evaluation, as Chen \cite{41}, we begin train our proposed MRHAM-YOLOv4-Slim model utilizing the PASCAL VOC \cite{42} dataset, which obtains pre-training weights containing abundant low-level image knowledge. Then, we initialize the weights employing the pre-trained weights when training the model using fetal US images. In order to demonstrate the generalization of our proposed model, we also adopt PASCAL VOC dataset to evaluate the MRHAM-YOLOv4-Slim. In this paper, we adopt four evaluation methods including recall: $R = N_{tp} / (N_{tp} + N_{fn})$, precision: $P = N_{tp} / (N_{tp} + N_{fp})$, F1 score: $F_1 = 2PR / (R + P)$ and mean average precision (mAP). The evaluation protocol of CIFAR-10 \cite{26} and CIFAR-100 \cite{26} is different from above methods, we employ Cumulative Matching Characteristic (CMC) curves to assess the various backbones.

\subsection{Implementation Details}
\textbf{Object Detection:} In fetal cardiac structures detection experiments, the training steps is 1,000 epochs and mini-batch size is set to 4. We utilize the Stochastic Gradient Descent (SGD) to optimize our proposed model. The initial learning rate and momentum is set to 0.0001 and 0.937, respectively. The step decay learning rate scheduling is cosine annealing strategy \cite{43}. L2 regularization method is adopted to clip weighting parameters and weight decay is set to 0.0005.

\textbf{Object Classification:} In CIFAR image classification experiments, mini-batch size is set to 256 to train various backbones on CIFAR-10 \cite{26} and CIFAR-100 \cite{26} datasets. The training steps is 3,00 epochs. The step decay learning rate adjusting strategy is used with initial learning rate 0.1 and multiply with a factor 0.1 at 150 epochs, 200 epochs and 250 epochs. We also adopt stochastic gradient descent (SGD) to optimize the different backbones, which the momentum and weight decay are set as 0.9 and 0.0005, respectively.
\subsection{Training the Networks}
Due to fetal US images data deficiencies, it is hard to obtain a significantly top model. To capture priori knowledge of images, we begin by training MRHAM-YOLOv4-Slim with PASCAL VOC 2012 \cite{42} dataset. The PASCAL VOC 2012 \cite{42} dataset consists of 17,125 images of 20 subjects. The MRHAM-YOLOv4-Slim is trained for 4,282,000 iterations using VOC images, obtaining pre-training weights with richly priori image knowledge. Then, we train the MRHAM-YOLOv4-Slim model for 250,000 iterations using fetal US images, initializing the model with the pre-trained weights. All the models proposed in this paper are implemented using Pytorch on a NVIDIA 2080 Ti GPU. In addition, all the speed data of FPS is calculated on the NVIDIA 2080Ti GPU.
\subsection{Fetal Four Cardiac Chambers Localization Performance}
In this subsection, we begin by analyzing the classification performance of different arrangements of the RCA module and RSA module. Then, we show the classification performance of two backbones with two attention modules (e.g., MRHAM and CBAM) on two public datasets. Finally, we show the performance of our proposed MRHAM-YOLOv4-Slim model. Furthermore, we compare the performance with the current state-of-the-art methods for locating fetal cardiac structures in US images.

\textbf{Arrangement of the RCA and RSA:} In this experiment, we compare five different ways of scheduling the RCA and RSA modules: single RCA, single RSA, sequential RCA-RSA, sequential RSA-RCA, and parallel use of both attention modules. Due to each module has a different role, the arranging strategy may affect the overall performance. The RCA is globally applied to the model, which focuses on the content context of targets. In addition, the RSA helps the model highlight the local location of targets. Hence, we perform extensive set of experiments to show the effect of various combinations of attention modules.

In Table I, we show quantitative results from our CSPDarknet53 baseline model when various combinations of the RCA and RSA are used to train the network on CIFAR-10. From the results, we can observe that all the combinations of RCA and RSA outperforms only using the RCA or RSA independently. Furthermore, an attention map generated by RCA and RSA in sequential order is finer than doing in parallel. In addition, we also find that the RCA-first order outperforms slightly better than the RSA-first order. This is consistent with the literature \cite{37}. Hence, we arrange the RCA and the RSA sequentially to form the MRHAM together.

\begin{table}[!htbp]
\label{1}\caption{Effect of different combining methods of the RCA and RSA modules}
\centering
\begin{tabular}{ccc}
\hline
\multicolumn{3}{c}{CIFAR-10}\\
\hline
Model&   Top-1&   Top-5\\
\hline
CSPDarknet53-RCA & 92.11& 99.74\\
CSPDarknet53-RSA & 92.16& 99.76\\
CSPDarknet53-(RCA and RSA in parallel)   & 92.19& 99.79\\
CSPDarknet53-(RSA+RCA) & 92.23& 99.81\\
CSPDarknet53-(RCA+RSA)& \textbf{92.36}& \textbf{99.86}\\

\hline
CSPDarknet53-Slim-RCA & 91.33& 99.74\\
CSPDarknet53-Slim-RSA & 91.36& 99.78\\
CSPDarknet53-Slim-(RCA and RSA in parallel) & 91.86& 99.82\\
CSPDarknet53-Slim-(RSA+RCA) & 92.40& 99.85\\
CSPDarknet53-Slim-(RCA+RSA)& \textbf{92.58}& \textbf{99.88}\\
\hline
\end{tabular}
\end{table}

\textbf{Effect of MRHAM:} To roundly evaluate the MRHAM, we perform extensive classification experiments on CIFAR-10 \cite{26} and CIFAR-100 \cite{26}. We evaluate our module in two networks including CSPDarknet53 and CSPDarknet53-Slim. In particular, CSPDarknet53-Slim is a 51-layers convolutional neural network, which is shallower than CSPDarknet53, a 67-layers convolutional architecture. As described in Section 4.2, CIFAR-10 includes 6,000 images per class yet CIFAR-100 has only 600 images per class.

Table II shows the experimental results. We observe that the two backbones with MRHAM perform better than their baselines considerably, which demonstrates the MRHAM can generalize well on CSPDarknet models in the different-scale dataset. Furthermore, the two CSPDarknets with MRHAM slightly improve the accuracy than that with CBAM \cite{37} which adopts pooling method to generate a descriptor. The CSPDarknet53 and CSPDarknet53-Slim are two finer backbones which abandon pooling method in the architecture. The CSPDarknets perform on par with state-of-the-art classifiers but with fewer floating point operations and more speed. In order to follow its style, the convolutional operation is mainly adopted to construct MRHAM architecture. Hence, the CSPDarknets with MRHAM yield a better performance. As shown in Table III, the MRHAM-YOLOv4-Slim achieves a higher measured frames per second (FPS) than CBAM-YOLOv4-Slim.

\begin{table}[!htbp]
\label{2}\caption{Effect of attention modules on different backbones}
\centering
\begin{tabular}{ccc}
\hline
\multicolumn{3}{c}{CIFAR-10}\\
\hline
Model&   Top-1&   Top-5\\
\hline
CSPDarknet53 & 92.01& 99.74\\
CBAM-CSPDarknet53 & 92.24& 99.79\\
MRHAM-CSPDarknet53& \textbf{92.36}& \textbf{99.86}\\

\hline
CSPDarknet53-Slim & 91.31& 99.70\\
CBAM-CSPDarknet53-Slim& 92.45& 99.84\\
MRHAM-CSPDarknet53-Slim& \textbf{92.58}& \textbf{99.88}\\
\hline
\multicolumn{3}{c}{CIFAR-100}\\
\hline
CSPDarknet53& 70.33& 90.87\\
CBAM-CSPDarknet53& 70.43& 91.12\\
MRHAM-CSPDarknet53& \textbf{70.75}& \textbf{91.49}\\
\hline
CSPDarknet53-Slim& 71.19& 91.26\\
CBAM-CSPDarknet53-Slim& 71.21& 91.36\\
MRHAM-CSPDarknet53-Slim& \textbf{72.37}& \textbf{91.99}\\
\hline
\end{tabular}
\end{table}

\textbf{MRHAM-YOLOv4-Slim Performance:} In order to clearly evaluate the effect of YOLOv4-Slim integrated with the MRHAM, we present several detailed statistics. Table III shows that the MRHAM-YOLOv4-Slim achieves the best performance on detecting fetal four cardiac chambers. In Table III, the YOLOv4-Slim embedded in the MRHAM module results in further performance gains. Furthermore, the MRHAM-YOLOv4-Slim has a better performance than YOLOv4, which demonstrates that our method is more effective. In addition, we can also observe that although MRHAM-YOLOv4-Slim performs slightly better than CBAM-YOLOv4-Slim, MRHAM-YOLOv4-Slim obtains higher FPS. This has great significance for real-time detection.

\begin{table}[!htbp]
\label{3}\caption{Comparison results of various YOLOv4 on fetal four cardiac chambers.}
\centering
\begin{tabular}{cccccc}
\hline
Model&   P&   R& F1 &mAP@0.5 &FPS\\
\hline
YOLOv4-Slim& 0.736& 0.786 & 0.76 & 0.752&\textbf{80}\\
YOLOv4& 0.798& 0.86 & 0.828 & 0.827&74\\
CBAM-YOLOv4-Slim& 0.837& 0.93 & 0.881 & 0.897&30\\
MRHAM-YOLOv4-Slim& \textbf{0.919}& \textbf{0.971} & \textbf{0.944} & \textbf{0.953} & 43\\
\hline
\end{tabular}
\end{table}

Fig. 5 gives visual results of these YOLOv4 networks in detecting four fetal cardiac chambers. From Fig. 5, we can see that the MRHAM-YOLOv4-Slim has the best accuracy in detection task.

Table IV compares the performance of our proposed MRHAM-YOLOv4-Slim against the current state-of-the-art methods on detecting fetal cardiac structures in US images. We observed that MRHAM-YOLOv4-Slim achieves the best performance among all the models, demonstrating the effectiveness of our proposed model in detecting four cardiac structures in FC plane. As illustrated in Table IV, the MRHAM-YOLOv4-Slim achieves the state-of-the-art mAP of $95.3\%$, while maintaining a comparable testing speed of 43 FPS. In addition, although the mAP of our proposed network is slightly higher than the Faster R-CNN (a two-stage network), the speed is considerably faster than Faster R-CNN.

\begin{figure}[!htbp]
	\centering
		\includegraphics[width=0.49\textwidth]{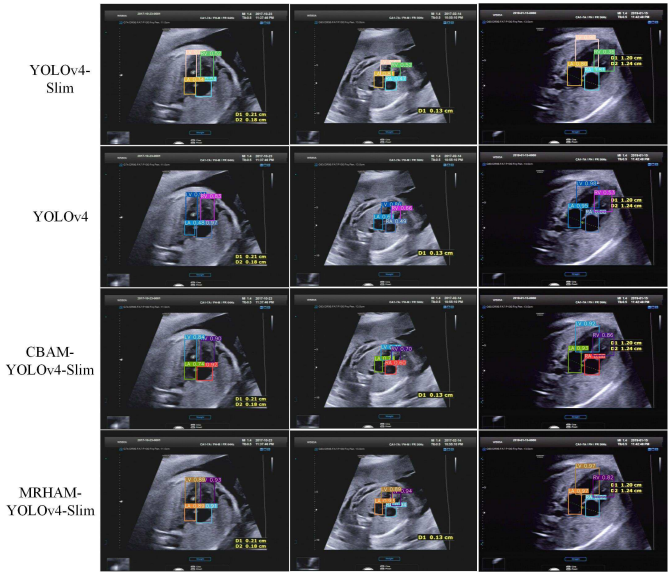}
	\caption{Visualization results in fetal cardiac anatomical structures detection. The same US images are adopted to test the detection performance of different networks.}
	\label{FIG:5}
\end{figure}

\begin{table}[!htbp]
\label{2}\caption{Comparison with the state-of-the-art methods}
\centering
\begin{tabular}{cccccc}
\hline
Model&   P&   R& F1 &mAP@0.5&FPS\\
\hline
SSD300 \cite{47}& 0.632& 0.721 & 0.674 & 0.65 & \textbf{88}\\
YOLOv4 \cite{27}& 0.798& 0.86 & 0.828 & 0.827&74\\
ARVBNet \cite{46}& 0.739& 0.878 & 0.801 & 0.832&81\\
YOLOv3 \cite{32}& 0.732& 0.896 & 0.805 & 0.853 & 83\\
YOLOv3-SPP \cite{32}& 0.744& 0.903 & 0.816 & 0.868 &80\\
Faster R-CNN \cite{30}& 0.747& 0.884 & 0.809 & 0.875 &8\\

MRHAM-YOLOv4-Slim& \textbf{0.919}& \textbf{0.971} & \textbf{0.944} & \textbf{0.953} & 43\\
\hline
\end{tabular}
\end{table}

\section{Conclusion and Future works}
In this paper, we propose a MRHAM-YOLOv4-Slim model to accurately locate fetal four important cardiac chambers of LV, LA, RV and RA in the FC views. The CSPDarknet53 as a CNN backbone of YOLOv4 for extracting discriminative features. Due to fetal US images data deficiencies, we simplify the CSPDarknet53 backbone named CSPDarknet53-Slim for alleviating the over-fitting. We use CSPDarknet53-Slim as the backbone architecture of our proposed MRHAM-YOLOv4-Slim. To improve the robustness and discrimination of features extracted by CSPDarknet53, we propose a MRHAM module. The original residual identity mapping is replaced by MRHAM in CSPDarknet53-Slim, capturing significantly discriminative features of fetal four cardiac chambers. Through considerable set of experiments, we show that our proposed MRHAM-YOLOv4-Slim model outperforms current state-of-the-art methods.

The detection of fetal four important cardiac chambers is the first step in our research of CHDs. Particularly, the pulmonary atresia with intact ventricular septum associated with hypoplastic right heart syndrome (PAIVSHRHS) is one of the intractable fetal CHDs in China. In the future, we will conduct a series of researches on PAIVSHRHS. Next, we will design an automatic classification fetal cardiac model. The classification model will automatically diagnose whether the fetal heart has PAIVSHRHS during obstetric examination, according to the fetal cardiac region of interest extracted by MRHAM-YOLOv4-Slim. It will early predict the growing status of fetal heart by automatic classification model, which is of great importance to families and society.

%\section{References}

\bibliography{mybibfile}

\end{document}